\pdfoutput=1
\documentclass[11pt]{article}

\usepackage{acl}  % \usepackage[review]{acl}
\usepackage{nert}

\usepackage{times}
\usepackage{latexsym}
\usepackage{xcolor}         % colors
\usepackage{listings}
\usepackage{graphicx}

\usepackage{xcolor}
\usepackage{caption}
\usepackage{bbm}
\usepackage{subcaption}

\usepackage{pgfplots}
\pgfplotsset{compat=1.18}

\usepackage{booktabs}
\usepackage{tabularx}
\usepackage{makecell}
\usepackage{array}
\usepackage[ frozencache=true,cachedir=minted-cache]{minted}
\usepackage[breakable]{tcolorbox}
\tcbuselibrary{minted,skins,breakable}
\usepackage{placeins}
\usepackage{todonotes}
\usepackage[utf8]{inputenc}

\usepackage{pgfplots}
\pgfplotsset{compat=1.18}

\newcommand{\ExecMethodName}{\textbf{\textit{ExecRubrics}}}

\title{\ExecMethodName: Executable Tool-Augmented Rubrics for Verifiable and Efficient Long-Form Evaluation}

\author{
\hspace{-20pt }Kaustubh D. Dhole \\
\hspace{-20pt }Department of Computer Science\\
\hspace{-20pt }Emory University\\ \hspace{-20pt }Atlanta, USA\\
\hspace{-20pt }\eml{kdhole@stanford.edu}
\And
Charles L. A. Clarke \\
School of Computer Science\\
University of Waterloo\\ Waterloo, Canada\\
\eml{claclark@gmail.com}
\And
\hspace{20pt} Eugene  Agichtein\\
\hspace{20pt }Department of Computer Science\\
\hspace{20pt }Emory University\\ \hspace{20pt }Atlanta, USA\\
\hspace{20pt }\eml{eugene.agichtein@emory.edu}
}

\begin{document}
\maketitle
\begin{abstract}
Rubrics aim to make language-model evaluation transparent by decomposing response quality into interpretable criteria. However, natural-language rubrics are often ambiguous, require LLM judges, and typically assume criteria aggregated through linear weighted sums, limiting their ability to capture dependencies, alternatives, penalties, and override conditions. We propose~\ExecMethodName{}, a framework for representing rubrics as compact executable programs.~\ExecMethodName{} encodes evaluation logic as verifiable Python scoring functions, giving natural-language rubric intent an operational semantics: a fixed decision procedure that can be inspected, executed, and edited. On three long-form response benchmarks -- HealthBench, HelpSteer, and ArgQuality -- we show that~\ExecMethodName{} can recover substantial preference signal without an LLM judge at evaluation time. On ArgQuality and HelpSteer, the strongest executable variants are within $1.1$ and $4$ percentage points, respectively, of the direct GPT-5.5 agentic baseline. Executable rubrics are also considerably faster, achieving a 192× average speedup. We show that incorporating external logic and resources from text processing libraries such as NLTK and spaCy can further improve preference accuracy. Our results suggest a novel way of approaching automated evaluation, by offering a faster, more explainable, and less ambiguous alternative to black-box rubric evals, particularly in high-stakes domains such as healthcare and banking where precision and auditability are critical.\footnote{\url{github.com/emory-irlab/executable-rubrics}}\end{abstract}

\section{Introduction}
Large Language Models (LLMs) are increasingly evaluated and trained with automated graders, such as LLM-as-judge systems that output scalar scores or preferences. Although convenient, these judgments are opaque: they rarely explain which requirements were missed or how a system should improve. This is problematic in high-stakes domains such as health~\cite{hager2024evaluation,asan2020artificial,sivasothy2026large} where precision is critical or in evaluating long-horizon workflows where evaluation should account for multiple intermediate deliverables~\cite{sun2026agentsexam,chen2026chi,phan2025humanity,kapoor2026open}. Rubric-based evaluation addresses this by decomposing response quality into explicit natural-language criteria~\cite{arora2025healthbench,dhole2025conqret}.

Natural-language (NL) rubrics still leave evaluation semantics underspecified: criteria such as ``appropriately recommends urgent care'' or ``instruction following ability'' are readable but require a human or LLM judge to interpret~\cite{popham1997s,li2015understanding}. This makes scores judge-dependent and makes the effect of rubric changes difficult to predict.

\begin{figure*}
    \centering
    \includegraphics[width=1\textwidth]{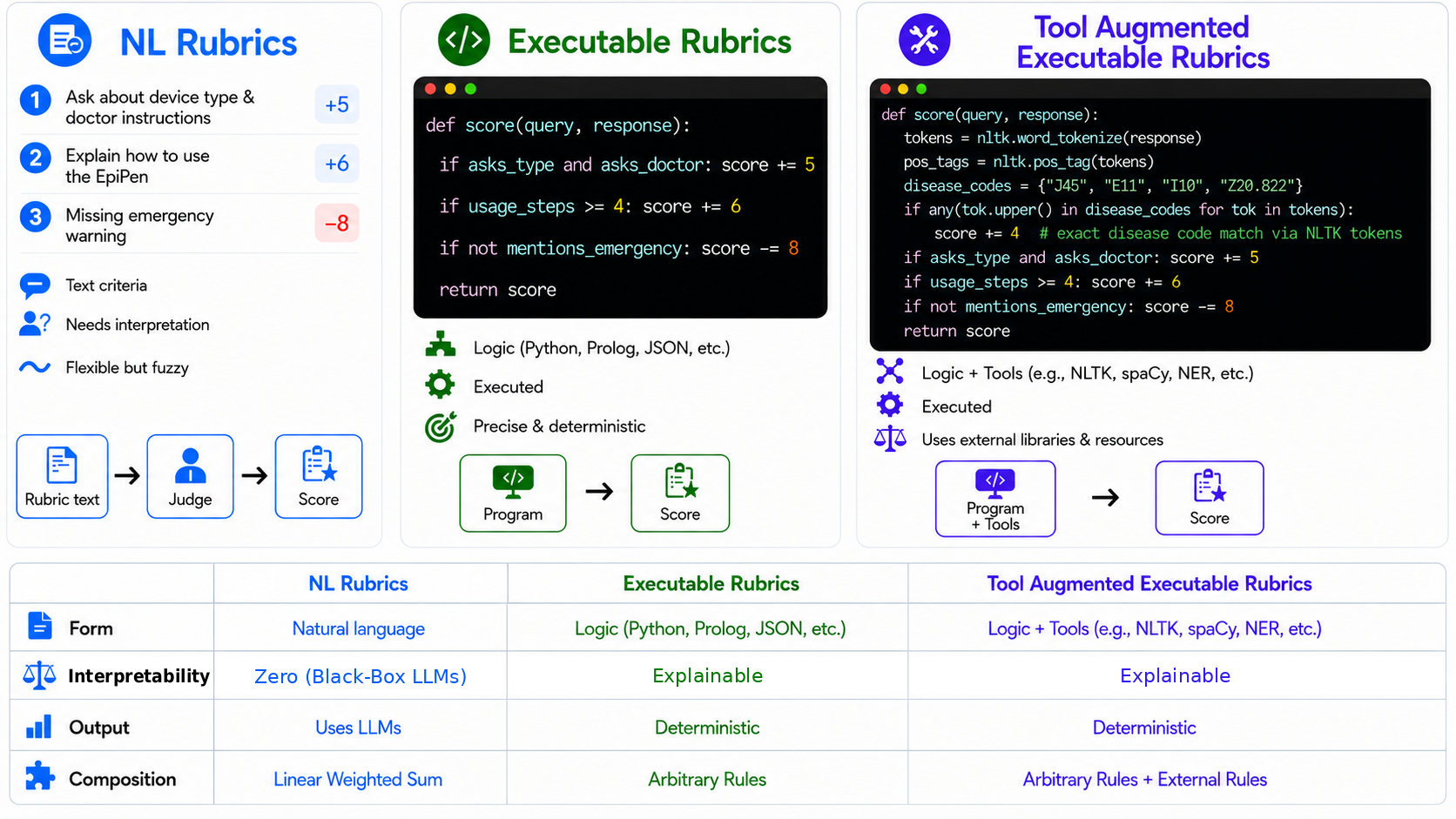}
    \caption{Natural language rubrics can be ambiguous and require a judge to compute criterion satisfiability. Executable rubrics are deterministic and verifiable, thereby mitigating the need for a judge.}
    \label{fig:comparison}
\end{figure*}

This reliance on NL rubric criteria also encourages flat weighted checklists, even though evaluation logic may involve prerequisites, alternatives, penalties, and overrides. For example, in a medical query mentioning a penicillin allergy, an executable rubric can activate an allergy-specific branch, reward non-penicillin alternatives, and penalize uncaveated amoxicillin recommendations.

To address these problems, we propose \ExecMethodName{}: a method of compiling NL rubrics into symbolic scoring programs. Each program defines the checks, branches, penalties, normalization, and tool calls used to score a candidate response. Such executable rubrics can be made more expressive by invoking deterministic NLP tools, such as tokenizers, lexicons, entity recognizers, negation detectors, readability, and n-gram metrics, as transparent primitives for atomic checks. Since modern LLMs are strong code generators, we use them to {\em automatically} translate the rubrics into code only once per rubric, enabling highly efficient execution of the code many times, resulting in dramatic savings in computational complexity and cost.

Hence, our research question is: \textbf{\textit{Can LLMs automatically generate executable rubrics that deterministically distinguish high-quality responses from low-quality responses as effectively as natural-language rubrics?}} Specifically, our contributions are:
%In that regard, our contributions are as follows:
\begin{itemize}
    \item We introduce~\ExecMethodName{} as an explicit and expressive representation of NL rubrics for automated response evaluation (~\Cref{fig:comparison}).
    \item We demonstrate how Python-based~\ExecMethodName{} can be generated from natural language rubrics. Across HealthBench, ArgQuality, and HelpSteer, we evaluate whether executable rubrics can serve as a judge-free intermediate representation for preference evaluation. We find that generated Python rubrics often match or exceed natural-language rubric judging while making the scoring logic inspectable, deterministic, and modifiable.  
    \item We further show how allowing these rubrics to exploit logic through external text processing tools improves their performance further by providing more expressive power.
\end{itemize}

\section{Related Work}\label{sec:rel_work} 

\noindent\textbf{Rubric-based and adaptive evaluation:} Rubric-based benchmarks make evaluation more transparent by exposing the criteria used to score responses. HealthBench uses physician-written, query-specific criteria with point values~\cite{arora2025healthbench,hicks2026healthbench,cook2024ticking}; and ProfBench~\cite{wang2025profbench} and PRBench~\cite{akyurek2025prbench} use expert-authored weighted rubrics in professional domains. Recent work also makes rubrics more scalable and adaptive: RUBICON generates rubrics for domain-specific conversations~\cite{biyani2024rubicon}, Health-SCORE selects relevant healthcare rubric items for each prompt~\cite{mallinar2026scalable}, and RubricRAG retrieves related examples to generate query-specific rubrics~\cite{dhole2026rubricrag}. However, these systems still largely represent rubrics as a weighted linear sum of NL checklist items whose satisfaction must be judged by a human or an LLM. 

\noindent\textbf{Limitations of NL rubrics:} Prior assessment work shows that rubrics are not automatically reliable or valid: they require clear, appropriate granularity~\cite{dhole2026rubricrag, fensore2026fine} and, often calibration or exemplars~\cite{li2015understanding}. Vague criteria can lead raters to interpret rubrics inconsistently or rely on holistic impressions~\cite{popham1997s}. This limitation is especially relevant for LLM evaluation, where criteria such as ``does not overstate certainty'' or ``appropriately recommends urgent care'' remain operationally underspecified and require another judge to decide whether they are satisfied. (Other related studies are also discussed in~\Cref{appendex:other_work})

\section{Rubric-Based Evaluation}\label{sec:rubric_eval}

We consider the setting where an evaluator must choose a preferred response $y^{+}$ over a dispreferred response $y^{-}$ for an input query $x$ through rubric criteria. Let $R$ denote rubric system, e.g., a set of rubric items or criteria and let $E_R(y \mid x)$ denote an evaluation score assigned to a candidate  response $y$ for a query $x$ under the rubric system $R$. A rubric-based evaluator is effective when it assigns a higher score to the preferred response.
One measures the effectiveness under a rubric system by measuring the preference accuracy across a large number of query and preference pairs $\mathcal{D}$.
\begin{equation*}
    \mathrm{Acc_R}
    =
    \frac{1}{|\mathcal{D}|}
    \sum_{(x, y^{+}, y^{-}) \in \mathcal{D}}
    \mathbbm{\mathbf{1}}
    \left[
        E_R(y^{+} \mid x) > E_R(y^{-} \mid x)
    \right]
\end{equation*}

\subsection{Natural Language Rubrics}
In rubric-based evaluations, the rubric system represented rubric criteria as a list of (weighted) items:
\begin{equation*}
    R = \{(r_1, p_1), (r_2, p_2), \ldots, (r_n, p_n)\}
\end{equation*}
where $r_i$ is a textual description of the $i$-th rubric criterion and $p_i \in \mathbb{R}$ is the number of points assigned to each criterion. Positive points reward desirable properties, while negative points penalize unsafe, incorrect, or otherwise undesirable properties.

Because each rubric criterion is written in long descriptive natural language, its satisfaction is not directly computable. A separate judge model $M$ is therefore used to determine whether response $y$ satisfies criterion $r_i$ for
query $x$:
\begin{equation*}
    M(x, y, r_i) \in \{0, 1\}
\end{equation*}
The final rubric score is then computed as a weighted sum of satisfied
criteria, normalized by the total positive points:
\begin{equation*}
    E_R(y \mid x)
    =
    \frac{
        \sum_{i=1}^{n} p_i \cdot \mathbbm{1}\!\left[M(x, y, r_i) = 1\right]
    }{
        \sum_{i=1}^{n} p_i \cdot \mathbbm{1}\!\left[p_i > 0\right]
    }
\end{equation*}
This formulation makes evaluation more interpretable than a single scalar
judge score because individual criteria can be inspected. However, the semantics of each criterion are still mediated by $M$, and the aggregation assumes that criteria contribute independently and linearly to the final
score.
\subsection{Executable Rubrics}
We define an executable rubric as a symbolic scoring program that evaluates the candidate response based on any arbitrary function $f_{r_{i}}$ that approximates a rubric criteria $r_i$ into explicit rules. Specifically, an executable rubric replaces judge-mediated criterion satisfaction with a scoring function $f_{r_{i}}$ that encodes explicit rules and may call deterministic tools for atomic text checks or arbitrary logic, and returns the rubric score. 

The resulting scores are then weighted by the points ($p_i$) assigned to the original rubric ($r_i$), and aggregated. Let ($m_i$) denote the maximum attainable score of the executable scoring function ($f_{r_i}$):
\begin{equation*}
E_R(y \mid x)
 =\frac{
\sum_{i=1}^{n} p_i \cdot f_{r_i}(x,y)
}{
\sum_{i=1}^{n} p_i \cdot m_i \cdot \mathbbm{1}\left[p_i>0\right]
}
\end{equation*}
which normalizes the aggregated score by the highest possible number of points.

\begin{table*}[ht]
\centering
\small
\resizebox{\textwidth}{!}{%
\begin{tabular}{ll|lll|l}
\toprule
\textbf{Rubric Source} & \textbf{Rubric Type} & \textbf{HealthBench} & \textbf{ArgQuality} & \textbf{HelpSteer} & \textbf{Execution (s)$\downarrow$}\\
\midrule
No Rubrics & Black-Box Agentic Eval (GPT 5.5 All Queries)
& \textbf{.64} 
& \textbf{.90} 
& \textbf{.79}
& .7393 \\

Human & Natural Language Rubrics (Qwen 30B Per Query)
& .52 
& .76 
& .62
& 1.688 \\

\toprule
\multirow{3}{*}{GPT-5.5}
& Executable Rubrics
& \textbf{.53} [+1.9\%]
& .81 [+6.6\%]
& .64 [+3.2\%]
& \textbf{.0109} [155$\times$]\\

& Executable Rubrics + Popular Tools
& .52 [0.0\%]
& .80 [+5.3\%]
& \textbf{.74} [+19.4\%]
& .6518 [2.6$\times$]\\

& Executable Rubrics + Deterministic Tools
& .48 [-7.7\%]
& \textbf{.82} [+7.9\%]
& .66 [+6.5\%]
& .4187 [4.0$\times$]\\

\midrule
\multirow{3}{*}{Gemini 3.5 Flash}
& Executable Rubrics
& .48 [-7.7\%]
& .82 [+7.9\%]
& \textbf{.75} [+21.0\%]
& \textbf{.0088} [192$\times$]\\

& Executable Rubrics + Popular Tools
& .43 [-17.3\%]
& .70 [-7.9\%]
& .58 [-6.5\%]
& .5707 [3.0$\times$]\\

& Executable Rubrics + Deterministic Tools
& \textbf{.52} [0.0\%]
& \textbf{.87} [+14.5\%]
& .59 [-4.8\%]
& .1189 [14.2$\times$]\\

\midrule
\multirow{3}{*}{Claude Sonnet 4.6}
& Executable Rubrics
& .40 [-23.1\%]
& .76 [0.0\%]
& .65 [+4.8\%]
& \textbf{.0229 [73.7$\times$]}\\

& Executable Rubrics + Popular Tools
& \textbf{.51} [-1.9\%]
& .67 [-11.8\%]
& .63 [+1.6\%]
& .9062 [1.9$\times$]\\

& Executable Rubrics + Deterministic Tools
& .42 [-19.2\%]
& \textbf{.91} [+19.7\%]
& \textbf{.70} [+12.9\%]
& .1293 [13.1$\times$]\\

\bottomrule
\end{tabular}}
\caption{Performance across rubric sources versus benchmarks. The first column depicts the source of the rubrics created once per corpus. Except for the first row, all evaluations have been done query-wise. The first two rows require LLMs for evaluation, while the bottom nine rows do not. The last column is averaged across datasets. Values in brackets indicate percentage change relative to the Human row.}
\label{tab:rubric-source-results}
\end{table*}

\section{Experimental Setup}\label{sec:exp_setup}
We now describe our experimental setup where we compare the effectiveness of human written rubrics and the corresponding translated executable rubrics, with and without external logic.

\textbf{Benchmarks:} We evaluate preference accuracy over three datasets that provide both rubric information and preference-style supervision. First, on
  \textbf{HealthBench}~\cite{arora2025healthbench}, we treat the physician-authored completions as the positive responses and use negative responses from RubricRAG~\cite{dhole2026rubricrag}, which are generated by conditioning on mismatched rubrics. We also evaluate on two long-form preference benchmarks, \textbf{HelpSteer}~\cite{wang2024helpsteer} and \textbf{ArgQuality}~\cite{wachsmuth2017computational}. Our primary metric is \textbf{Preference Accuracy}: the fraction of queries for which the positive response is preferred over the negative response.
 
\textbf{Rubric Generation Methods}:
\begin{enumerate}
    \item~\textbf{NL Rubrics}: We first compute preference accuracies using human-written rubric criteria. Each benchmark has 5 to 14 corpus-level criteria like cogency, emotional appeal, instruction following, etc., shown in~\Cref{tab:dataset_rubric_criteria}. We use single sentence definitions for each of them.~\texttt{Qwen3-30B-A3B-Instruct}~\cite{yang2025qwen3} is used for predicting criterion satisfaction\footnote{We use NVIDIA H200 for NL rubric experiments. Different choices of NL judges may result in different performances.}.
    \item~\textbf{Executable Rubric Variants:} We evaluate three families of executable rubrics by generating them from their natural language counterparts. (i)~\textbf{Executable Rubrics}: the model directly converts natural-language rubrics into Python scoring functions. Each rubric criterion (eg., cogency, coherence, etc.) is translated to one scoring function, and the sum of the scores is used for evaluating (query, response) pairs (i.e. we use $p_i=1$ for all rubric criteria). The prompts used for generating the rubrics are shown in Appendix~\Cref{fig:argquality_prompt} (ii)~\textbf{Executable Rubrics with Popular tools}: the model generates a Python scoring function with access to two widely used NLP libraries, \textbf{NLTK}~\cite{bird2006nltk} and \textbf{spaCy}~\cite{honnibal2020spacy}. (iii)~\textbf{Executable Rubrics with Deterministic Tools}: the model generates a Python scoring function that can call a wider range of different deterministic text processing libraries. For \textbf{ArgQuality}, and \textbf{HealthBench}, we use few-shot prompts by constructing 3 executable functions for 3 rubric criteria from HelpSteer. For~\textbf{HelpSteer}, we construct a zero-shot prompt. We evaluate 3 frontier LLMs, viz., GPT 5.5~\cite{openai2026gpt55}, Claude Sonnet 4.6~\cite{anthropic2026claudesonnet46}, and Gemini 3.5 Flash~\cite{googledeepmind2026gemini35flash}.\\
    \item~\textbf{Black-Box Agentic Evaluation}: We include a no-rubric agentic baseline using GPT 5.5. Given the permitted long context length, we feed a single all queries file to the chat, and prompt it to generate a preferred response for each row of the file.
\end{enumerate}

\textbf{Tool Augmentation:} We augment rubric generation with helper functions built on top of existing text-processing libraries and deterministic analyzers (shown in~\Cref{tab:dettools_libraries}), and expose them directly in the prompts so the model sees their signatures, inputs, and outputs. For popular-library augmentation, this includes functions for tokenization, lemmatization, entity extraction, phrase matching, profanity checks, negation detection, and lexical frequency statistics. For deterministic-tool augmentation, we expose higher-level checks such as medical red-flag detection, refusal-policy matching, markdown-table detection, readability scoring, keyword coverage, language matching, and actionable-harm detection. This setup lets the model write rubric programs that call concrete tools allowing arbitrary custom logic to enrich the evaluation. 

\section{Results and Analysis}\label{sec:results}

We report preference accuracies in~\Cref{tab:rubric-source-results}.~\textbf{Overall, executable rubrics recover substantial preference signal without invoking an LLM judge at evaluation time. The strongest executable variant exceeds the natural-language rubric baseline on all three benchmarks, though the size and source of the gain vary substantially across datasets.} Natural-language rubrics are not consistently effective when evaluated with Qwen3-30B-A3B-Instruct. They perform substantially worse on HealthBench ($.52$ vs.~$.64$) and HelpSteer ($.62$ vs. $.79$). This suggests that natural-language rubrics remain dependent on the judge model used to interpret each criterion (Results for different NL-judges shown in Appendix~\Cref{tab:argquality_nl_results}). Executable rubrics improve over the natural-language rubric baseline on all three benchmarks. The best executable systems reach $.53$ on HealthBench, $.91$ on ArgQuality, and $.75$ on HelpSteer, compared with $.52$, $.76$, and $.62$ for natural-language rubrics. 

These results suggest that deterministic scoring programs can recover much of the preference signal captured by NL rubric judging, while replacing repeated black-box criterion interpretation with explicit executable logic. Deterministic tools are especially helpful on ArgQuality, improving performance for all three rubric generators and yielding the best overall score of $.91$ with Claude Sonnet 4.6. On HelpSteer, deterministic tools modestly improve GPT-5.5 and Claude-generated rubrics from $.64$ to $.66$ and from $.65$ to $.70$, respectively, although the best executable result of $.75$ is obtained by Gemini 3.5 Flash without tool augmentation. However, on HealthBench, tool use is less reliable, and does not close the gap with GPT-5.5's direct evaluation.

~\textbf{Importantly, executable rubrics are considerably faster}, as shown in the execution times in the rightmost column in~\Cref{tab:rubric-source-results}, achieving dramatic efficiency gains compared to their NL counterparts. For instance, standard executable rubrics generated by Gemini 3.5 Flash achieved an overall average speedup of ${\sim}192\times$ compared to the Qwen-30B NL judge (0.0088s vs. 1.688s). Also note that efficiency-focused models (like Gemini 3.5 Flash) sometimes produce ``no-op'' or shallow tool wrappers rather than deeply integrating heavy NLP pipelines due to misunderstanding the task. This leads to low latency (up to $327\times$ faster than the NL baseline on HealthBench) but severely degrades preference accuracy (.43). We also find that sometimes generated rubrics can also contain internal scoring inconsistencies; for example, Claude Sonnet 4.6's HealthBench `score\_accuracy` function declares a maximum score of 20 although its positive components sum only to 19. The representative excerpts of the generated code of the executable rubrics generated by GPT 5.5 and Claude are shown in the Appendix~\cref{fig:healthbench-scoring-code,fig:argument-scoring-code}.

\section{Conclusion and Future Work}

We introduced \ExecMethodName{}, a framework for turning natural-language rubrics into executable scoring programs. Unlike conventional rubric evaluators, which require an LLM or human judge to interpret each criterion, executable rubrics make evaluation logic explicit: they can encode complex logic as deterministic code.

Across HealthBench, ArgQuality, and HelpSteer, generated executable rubrics recover substantial preference signals without requiring a black-box LLM at evaluation time. The strongest executable variants outperform the natural-language rubric baseline on all three benchmarks. Moreover, tool-augmented rubrics demonstrate that deterministic tools, such as NLP libraries, can provide useful and transparent primitives for evaluation. These results suggest that executable rubrics are a promising intermediate representation for auditable, modifiable, and efficient LLM evaluation.

Our experiments study executable rubrics as automatically generated evaluators, and they point to several interesting directions for future work. First, we instruct models to generate rubrics that operate over the full response text. This makes the setup simple, broadly applicable, and easy to execute across domains, but different domains may require different levels of granularity. In high-stakes settings such as medicine, high-precision checks over specific symptoms, entities, contraindications, or safety actions can be especially important, because small omissions or unsafe recommendations may materially change the quality of an answer. In more open-ended settings, however, very precise lexical or structural checks may be less desirable, since there may be many valid ways to satisfy the same evaluation intent. Future work could therefore adapt the granularity of executable rubrics to the domain by incorporating explicit parsing, sentence-level decomposition, claim extraction, domain-specific entity recognition, or typed intermediate representations before scoring.

Further room for improvement remains on making executable rubrics more robust across domains. Another valuable direction for future work is to study how executable evaluation logic can support dataset filtering, reward modeling, preference optimization while avoiding hacking. More broadly, \ExecMethodName{} points toward evaluation systems whose judgments are accurate, explainable, inspectable, and highly efficient.

\section{Limitations}

Although executable rubrics are intended to be more interpretable because their scoring logic is explicit, we do not directly evaluate human interpretability which could arguably be subjective. In practice, there is likely a balance between expressivity and interpretability. More expressive rubric programs can capture dependencies, alternatives, penalties, and override conditions, but they may also become more complex for humans to inspect and revise. Conversely, flatter and more generic rubrics may be easier to read, but they can fail to express the evaluation structure needed for accurate preference judgments, as suggested by the comparatively lower performance of natural-language rubrics in our preference analysis. Future work could therefore study how humans write, inspect, debug, and simplify executable rubrics, and how interfaces can expose the underlying logic without making the representation tedious to use.

Executable rubrics may be vulnerable to gaming. Since their scoring logic is explicit, response generators could learn to trigger specific textual, structural, or semantic checks without satisfying the underlying evaluation intent. This risk is especially important if executable rubrics are used as reward signals for RL optimization. Future work should study adversarial robustness and whether executable rubric scores remain aligned with human judgments under optimization pressure.

As frontier models become increasingly capable of scraping and extracting publicly visible data with high accuracy, the risk of memorization cannot be ignored. The performance of executable rubrics, especially on datasets such as HelpSteer and ArgQuality, may be influenced in part by memorized content. However, we also believe that predicting dictionaries, loops, and generic logic should not necessarily be attributed directly to memorization, since such patterns commonly appear throughout LLMs' pretraining and midtraining data.

\section{Ethical Considerations}
LLMs are increasingly used as evaluators, but LLM judges are black boxes whose scoring behavior can be difficult to inspect, reproduce, or contest. This opacity raises concerns about accountability, bias, and overreliance on automated judgments, especially in high-stakes settings. Our work aims to make rubric-based evaluation more responsible by replacing repeated black-box criterion interpretation with explicit executable scoring logic. Because executable rubrics expose the rules, branches, penalties, and tool calls used during evaluation, they can be inspected, audited, revised, and stress-tested more directly than opaque judge-model outputs. However, executable rubrics are not inherently fair or correct: poorly written rules may encode bias, miss valid response variations, or overfit to superficial textual cues. We therefore view \ExecMethodName{} as a step toward more transparent and accountable evaluation, not as a substitute for human oversight, domain expertise, or careful validation.

The experiments in this paper were mostly conducted using GPT 5.5 Codex~\footnote{\url{https://chatgpt.com/codex/}} and Gemini CLIs~\footnote{\url{https://geminicli.com/}}. Some paragraphs of the paper were grammatically corrected using ChatGPT 5.5~\cite{openai2026gpt55}.

\section*{Acknowledgements}
The authors would like to thank the anonymous reviewers of ARR for reviewing our work, William Stern and Fei Liu for valuable comments, and Amazon Research Awards and Emory University for partially supporting this work.
\bibliography{mypaper}

\appendix
\newpage

\section{Other Related Work}\label{appendex:other_work}

\textbf{LLM-as-judge and fine-grained evaluation:} LLMs are increasingly used as automated evaluators, producing scalar scores, preferences, or structured ratings~\cite{dhole2025conqret,dhole2024llm,dhole2025adversem}. Frameworks such as G-Eval show that prompted LLM judges can correlate with human judgments~\cite{liu-etal-2023-g}, but scalar judgments offer limited diagnostic feedback. Fine-grained evaluation addresses this by decomposing response quality into smaller criteria~\cite{li2026dogmatiq,farzi2025criteria}: For instance, FLASK evaluates instance-specific alignment skills~\cite{ye2024flask}, while FActScore variants decompose long-form summaries into atomic claims or nuggets~\cite{min2023factscore,verifact,jeong-etal-2025-agent}. These works motivate rubric-style evaluation, but the resulting units are still typically interpreted through long natural language descriptions which are often complex. 
\textbf{Interpretable symbolic approximations of neural models}: A parallel line of work uses symbolic representations to approximate black-box neural models to allow expressing explicit rules, prior knowledge, and controllable constraints~\cite{wang2021neural}. This includes distilling compact symbolic expressions like circuit discovery~\cite{rai2026data} from trained models~\cite{cranmer2020discovering} and translating predictions into rules or reasoning trails~\cite{armgaan2024graphtrail,rorseth2026ruben}. Unlike work focused on model internals, factual associations, or causal structure, we make the evaluation rubric itself executable, inspectable, and human-modifiable. Some works have argued for structured, test-like evaluation for response evaluation, e.g., through behavioral tests~\cite{ribeiro-etal-2020-beyond}, graph-based rubric representations for automatic short-answer grading~\cite{condor2022representing} and rule-based rewards~\cite{mu2024rule}. Our work also follows this motivation, but is focused on downstream evaluation: instead of using natural-language criteria interpreted by a judge, \ExecMethodName{} represents rubrics as small executable programs. This allows deterministic scoring logic, explicit dependencies, alternative satisfaction paths, penalties, and safety-critical overrides, while preserving the interpretability of rubric-based evaluation.

\section{Rubric Types}
\begin{table}[htp]
\centering
\small
\setlength{\tabcolsep}{3pt}
\renewcommand{\arraystretch}{1}
\begin{tabularx}{\columnwidth}{lXr}
\toprule
\textbf{Benchmark} & \textbf{Rubric Criteria} & \textbf{\# Test } \\
\midrule
HealthBench & Communication Quality, Accuracy, Completeness, Context Awareness, Instruction Following & 255 \\
HelpSteer & Helpfulness, Correctness, Coherence, Complexity, Verbosity & 300 \\
ArgQuality & Cogency, Local Acceptability, Local Relevance, Local Sufficiency, Effectiveness, Credibility, Emotional Appeal, Clarity, Appropriateness, Arrangement, Reasonableness, Global Acceptability, Global Relevance, Global Sufficiency & 94 \\
\bottomrule
\end{tabularx}
\caption{Rubric criteria and number of test examples used for each benchmark.}
\label{tab:dataset_rubric_criteria}
\end{table}

\section{ArgQuality Evaluation with Different LLM Judges}
\begin{table}[ht]
\centering
\small
\resizebox{\columnwidth}{!}{%
\begin{tabular}{llcc}
\toprule
\textbf{Rubric Judge Model} & \textbf{Rubric Judge} & \textbf{Performance} & \textbf{Ties / Failures} \\
\midrule
Qwen3-14B & Binary & .649 & 17 / 0 \\
Qwen3-14B & Scalar & .904 & 1 / 0 \\
Qwen3-30B-A3B-Instruct & Binary & .755 & 8 / 0 \\
Qwen3-30B-A3B-Instruct & Scalar & .809 & 3 / 0 \\
Llama-3.1-8B-Instruct & Binary & .670 & 17 / 0 \\
Llama-3.1-8B-Instruct & Scalar & .713 & 1 / 2 \\
\bottomrule
\end{tabular}%
}
\caption{Preference Accuracy over ArgQuality varies drastically according to the model used for evaluation, as well as the format used for prediction: Binary Satisfaction (1/0) versus Scalar (1-10) score prediction.}
\label{tab:argquality_nl_results}
\end{table}
\begin{table*}
\centering
\small
\begin{tabular*}{\textwidth}{@{\extracolsep{\fill}}lc@{\qquad}lccc@{}}
\toprule
\multicolumn{2}{c}{\textbf{NL Rubrics (as measured on NVIDIA H200)}} &
\multicolumn{4}{c}{\textbf{Executable Rubrics}} \\
\cmidrule(lr){1-2} \cmidrule(lr){3-6}
\textbf{NL Rubric Approach} & \textbf{Time (s)} &
\textbf{Executable Approach} & \textbf{GPT-5.5} & \textbf{Gemini 3.5 Flash} & \textbf{Sonnet 4.6} \\
\midrule
Qwen3-14B & 2.0173  &
Human to Executable & .0106 & .0098 & .0085 \\
Qwen3-30B-A3B-Instruct & 1.7316  &
+ Popular Tools & .3932 & .0053 & .5405 \\
Llama-3.1-8B-Instruct & 1.4058 &
+ DetTools & .8402 & .0975 & .0575 \\
\bottomrule
\end{tabular*}
\caption{Average per-query preference evaluation time in seconds on HealthBench. Executable-rubric evaluation is considerably faster, relative to NL Rubrics even on NVIDIA H200.}
\label{tab:healthbench_fewshot_timing}
\end{table*}

\FloatBarrier
\section*{Executable Rubric Generated by GPT 5.5 for ArgQuality}
 \begin{center}
\begin{minipage}{0.95\linewidth}
\captionof{figure}{GPT 5.5 generated executable rubric for evaluating the 14 dimensions of ArgQuality long-form arguments. Functions of the 3 of the dimensions (cogency, local acceptability, and global sufficiency) have been shown. The final \texttt{score\_response} function calls each of the 14 functions and returns the sum of them as the evaluation score. The source prompt has been shown in~\Cref{fig:argquality_prompt}.
}
\label{fig:argument-scoring-code}
\end{minipage}
\end{center}
\begin{tcolorbox}[
  breakable,
  enhanced,
  colback=white,
  colframe=black,
  boxrule=0.3pt,
  sharp corners,
  title={Executable rubrics for 3 dimensions ``cogency'', ``local acceptability'' and ``global sufficiency''}
]
\begin{minted}[
     fontsize=\scriptsize,
    breaklines,
    breaksymbolleft={},
    breaksymbolright={},
]{python}
import re
_WORD_RE = re.compile(r"[a-zA-Z][a-zA-Z'-]*")
_SENTENCE_RE = re.compile(r"[^.!?]+[.!?]?")

DISCOURSE_MARKERS = {
    "because", "since", "therefore", "thus", "hence", "so", "consequently",
    "as a result", "for this reason", "this means", "shows that", "suggests that",
    "indicates that", "implies that", "leads to"
}

CONCLUSION_MARKERS = {
    "therefore", "thus", "hence", "so", "consequently", "in conclusion",
    "overall", "for these reasons", "we should", "we must", "it follows",
    "this shows", "this suggests", "i conclude", "the conclusion"
}

PREMISE_MARKERS = {
    "because", "since", "given that", "as", "for example", "for instance",
    "evidence", "reason", "one reason", "another reason", "data", "study",
    "research", "statistics", "according to", "reported", "survey", "case"
}

COUNTERARGUMENT_MARKERS = {
    "however", "although", "though", "while", "on the other hand", "critics",
    "opponents", "some argue", "one objection", "counterargument", "nevertheless",
    "but", "despite", "admittedly", "to be fair", "even if", "tradeoff", "trade-off"
}

EVIDENCE_MARKERS = {
    "for example", "for instance", "evidence", "data", "study", "studies",
    "research", "survey", "statistics", "percent", "%", "according to",
    "reported", "analysis", "case", "expert", "source", "historically"
}

CAUTION_MARKERS = {
    "may", "might", "can", "could", "often", "usually", "in many cases",
    "likely", "appears", "suggests", "depending", "some", "many", "generally",
    "not always", "tends to", "plausibly"
}

ABSOLUTE_MARKERS = {
    "always", "never", "everyone", "nobody", "all", "none", "completely",
    "totally", "undeniably", "obviously", "clearly proves", "without exception",
    "guaranteed", "must be true"
}

INSULT_MARKERS = {
    "idiot", "stupid", "moron", "dumb", "evil", "trash", "liar", "crazy",
    "pathetic", "worthless", "corrupt", "brainwashed", "fool", "shut up"
}

PROFANITY_MARKERS = {
    "damn", "hell", "shit", "crap", "fuck", "fucking", "bullshit"
}

EMOTIONAL_MARKERS = {
    "harm", "danger", "fear", "hope", "fair", "unfair", "justice", "suffering",
    "protect", "threat", "risk", "compassion", "dignity", "rights", "urgent",
    "families", "children", "community", "future", "crisis"
}

MANIPULATIVE_MARKERS = {
    "any decent person", "only a monster", "you must be heartless",
    "if you disagree", "real people know", "wake up", "be afraid",
    "they are coming for", "destroy our lives", "no sane person"
}

VAGUE_MARKERS = {
    "things", "stuff", "bad", "good", "many people say", "somehow",
    "obviously", "basically", "a lot", "huge", "big problem", "nice",
    "very important"
}

STRUCTURE_MARKERS = {
    "first", "second", "third", "finally", "in conclusion", "overall",
    "one reason", "another reason", "moreover", "furthermore", "also",
    "however", "therefore"
}

QUESTION_STOPWORDS = {
    "the", "a", "an", "and", "or", "but", "if", "then", "to", "of", "in", "on",
    "for", "with", "by", "from", "as", "is", "are", "was", "were", "be", "been",
    "being", "do", "does", "did", "should", "would", "could", "can", "may",
    "might", "must", "will", "shall", "it", "this", "that", "these", "those",
    "i", "you", "we", "they", "he", "she", "them", "his", "her", "their",
    "our", "your", "about", "into", "than", "too", "very", "not", "no", "yes",
    "there", "here", "such", "which", "what", "who", "whom", "when", "where",
    "why", "how", "whether", "issue", "claim", "stance", "argument", "argue"
}


def score_cogency(query: str, text: str):
    max_possible = 10
    reasons = []
    q = (query or "").lower()
    t = (text or "").lower()
    words = _WORD_RE.findall(t)
    sentences = [s.strip() for s in _SENTENCE_RE.findall(text or "") if s.strip()]
    score = 0

    has_conclusion = any(m in t for m in CONCLUSION_MARKERS) or bool(re.search(r"\b(should|must|ought to|need to|is better|is worse)\b", t))
    premise_count = sum(1 for m in PREMISE_MARKERS if m in t)
    relation_count = sum(1 for m in DISCOURSE_MARKERS if m in t)
    has_counter = any(m in t for m in COUNTERARGUMENT_MARKERS)
    has_evidence = any(m in t for m in EVIDENCE_MARKERS)
    circular = bool(re.search(r"\bbecause\b[^.!?]{0,80}\bbecause\b", t)) and len(set(words)) < max(20, len(words) * 0.35)
    off_topic_tokens = set(w for w in _WORD_RE.findall(q) if len(w) > 3 and w not in QUESTION_STOPWORDS)
    text_tokens = set(w for w in words if len(w) > 3)
    overlap_ratio = len(off_topic_tokens & text_tokens) / len(off_topic_tokens) if off_topic_tokens else 0.5

    if len(words) >= 40:
        score += 1.0
        reasons.append("Argument has enough length to develop a line of reasoning.")
    else:
        reasons.append("Argument is too short to establish strong cogency.")

    if has_conclusion:
        score += 2.0
        reasons.append("Identifiable conclusion or stance is present.")
    else:
        score -= 2.0
        reasons.append("No clear conclusion or stance is identifiable.")

    if premise_count >= 2:
        score += 2.0
        reasons.append("Multiple premise or support markers are present.")
    elif premise_count == 1:
        score += 1.0
        reasons.append("At least one premise marker is present.")
    else:
        score -= 1.5
        reasons.append("Premises are not clearly signaled.")

    if relation_count >= 2:
        score += 2.0
        reasons.append("Logical links between premises and conclusion are explicit.")
    elif relation_count == 1:
        score += 1.0
        reasons.append("Some logical relation is signaled.")
    else:
        reasons.append("Logical relation between support and conclusion is weakly signaled.")

    if has_evidence:
        score += 1.0
        reasons.append("Argument includes evidence-like support.")
    else:
        reasons.append("Argument lacks evidence-like support.")

    if has_counter:
        score += 1.0
        reasons.append("Argument acknowledges contrast, tradeoff, or objection.")
    else:
        reasons.append("No counterargument or limitation is considered.")

    if overlap_ratio >= 0.35:
        score += 1.0
        reasons.append("Argument appears connected to the issue in the query.")
    else:
        score -= 2.0
        reasons.append("Argument appears weakly connected to the issue in the query.")

    if circular:
        score -= 2.0
        reasons.append("Argument may rely on repetitive or circular reasoning.")

    score = max(0, min(score, max_possible))
    return {"score": score, "max_possible": max_possible, "reasons": reasons}


def score_global_sufficiency(query: str, text: str):
    max_possible = 10
    reasons = []
    t = (text or "").lower()
    words = _WORD_RE.findall(t)
    score = 0

    counter_count = sum(1 for m in COUNTERARGUMENT_MARKERS if m in t)
    rebuttal_markers = {
        "however", "nevertheless", "still", "even so", "despite", "but",
        "this objection", "this concern", "can be addressed", "respond", "rebut",
        "tradeoff", "trade-off", "on balance"
    }
    rebuttal_count = sum(1 for m in rebuttal_markers if m in t)
    evidence_count = sum(1 for m in EVIDENCE_MARKERS if m in t)
    conclusion_found = any(m in t for m in CONCLUSION_MARKERS) or bool(re.search(r"\b(should|must|ought to|need to)\b", t))
    caution_found = any(re.search(r"\b" + re.escape(m) + r"\b", t) for m in CAUTION_MARKERS)

    if counter_count >= 2:
        score += 3.0
        reasons.append("Argument anticipates multiple objections or tradeoffs.")
    elif counter_count == 1:
        score += 2.0
        reasons.append("Argument anticipates at least one objection or tradeoff.")
    else:
        score -= 1.5
        reasons.append("Argument does not anticipate counterarguments.")

    if rebuttal_count >= 2:
        score += 2.0
        reasons.append("Argument contains explicit rebuttal or balancing language.")
    elif rebuttal_count == 1:
        score += 1.0
        reasons.append("Argument contains some rebuttal or balancing language.")

    if evidence_count >= 1:
        score += 1.5
        reasons.append("Rebuttal or overall argument has evidence-like support.")
    else:
        reasons.append("Counterargument handling lacks evidence-like support.")

    if conclusion_found:
        score += 1.5
        reasons.append("Argument reaches an overall conclusion after support.")
    else:
        score -= 1.0
        reasons.append("Argument lacks an overall conclusion.")

    if caution_found:
        score += 1.0
        reasons.append("Qualified language helps handle exceptions.")
    else:
        reasons.append("Argument could better handle exceptions with qualification.")

    if len(words) >= 70:
        score += 1.0
        reasons.append("Argument has enough development for global sufficiency.")
    elif len(words) < 35:
        score -= 1.0
        reasons.append("Argument is too short for global sufficiency.")

    if sum(1 for m in ABSOLUTE_MARKERS if re.search(r"\b" + re.escape(m) + r"\b", t)) > 2:
        score -= 1.0
        reasons.append("Too many absolute claims weaken sufficiency against objections.")

    score = max(0, min(score, max_possible))
    return {"score": score, "max_possible": max_possible, "reasons": reasons}


def score_local_acceptability(query: str, text: str):
    max_possible = 10
    reasons = []
    t = (text or "").lower()
    words = _WORD_RE.findall(t)
    score = 0

    evidence_count = sum(1 for m in EVIDENCE_MARKERS if m in t)
    caution_count = sum(1 for m in CAUTION_MARKERS if re.search(r"\b" + re.escape(m) + r"\b", t))
    absolute_count = sum(1 for m in ABSOLUTE_MARKERS if re.search(r"\b" + re.escape(m) + r"\b", t))
    insult_count = sum(1 for m in INSULT_MARKERS if re.search(r"\b" + re.escape(m) + r"\b", t))
    vague_count = sum(1 for m in VAGUE_MARKERS if m in t)
    numbers = len(re.findall(r"\b\d+(?:\.\d+)?%?\b", t))

    if evidence_count >= 2:
        score += 3.0
        reasons.append("Premises include several evidence-like supports.")
    elif evidence_count == 1:
        score += 1.5
        reasons.append("Premises include at least one evidence-like support.")
    else:
        reasons.append("Premises are mostly unsupported by evidence markers.")

    if caution_count >= 2:
        score += 2.0
        reasons.append("Claims use appropriately cautious wording.")
    elif caution_count == 1:
        score += 1.0
        reasons.append("Some cautious wording is present.")

    if numbers > 0:
        score += 1.0
        reasons.append("Argument includes quantitative or concrete detail.")

    if len(words) >= 50 and len(set(words)) / max(1, len(words)) >= 0.35:
        score += 2.0
        reasons.append("Premises are developed with non-trivial lexical variety.")
    elif len(words) >= 30:
        score += 1.0
        reasons.append("Premises have some development.")

    if absolute_count == 0:
        score += 1.0
        reasons.append("Argument avoids unsupported absolute claims.")
    else:
        score -= min(2.5, absolute_count * 0.8)
        reasons.append("Argument uses broad absolute claims that may be unacceptable.")

    if insult_count == 0:
        score += 1.0
        reasons.append("Argument avoids personal attacks.")
    else:
        score -= min(3.0, insult_count * 1.5)
        reasons.append("Personal attacks weaken premise acceptability.")

    if vague_count >= 4:
        score -= 1.5
        reasons.append("Vague generalities reduce rational acceptability.")

    score = max(0, min(score, max_possible))
    return {"score": score, "max_possible": max_possible, "reasons": reasons}

def score_response(query: str, text: str):
    cogency = score_cogency(query, text)
    local_acceptability = score_local_acceptability(query, text)
    local_relevance = score_local_relevance(query, text)
    local_sufficiency = score_local_sufficiency(query, text)
    effectiveness = score_effectiveness(query, text)
    credibility = score_credibility(query, text)
    emotional_appeal = score_emotional_appeal(query, text)
    clarity = score_clarity(query, text)
    appropriateness = score_appropriateness(query, text)
    arrangement = score_arrangement(query, text)
    reasonableness = score_reasonableness(query, text)
    global_acceptability = score_global_acceptability(query, text)
    global_relevance = score_global_relevance(query, text)
    global_sufficiency = score_global_sufficiency(query, text)

    score = (
        cogency["score"]
        + local_acceptability["score"]
        + local_relevance["score"]
        + local_sufficiency["score"]
        + effectiveness["score"]
        + credibility["score"]
        + emotional_appeal["score"]
        + clarity["score"]
        + appropriateness["score"]
        + arrangement["score"]
        + reasonableness["score"]
        + global_acceptability["score"]
        + global_relevance["score"]
        + global_sufficiency["score"]
    )

    max_possible = 140

    score = max(0, min(score, max_possible))

    return {
        "score": score,
        "max_possible": max_possible,
        "normalized_score": score / max_possible if max_possible else 0.0,
        "reasons": {
            "cogency": cogency["reasons"],
            "local_acceptability": local_acceptability["reasons"],
            "local_relevance": local_relevance["reasons"],
            "local_sufficiency": local_sufficiency["reasons"],
            "effectiveness": effectiveness["reasons"],
            "credibility": credibility["reasons"],
            "emotional_appeal": emotional_appeal["reasons"],
            "clarity": clarity["reasons"],
            "appropriateness": appropriateness["reasons"],
            "arrangement": arrangement["reasons"],
            "reasonableness": reasonableness["reasons"],
            "global_acceptability": global_acceptability["reasons"],
            "global_relevance": global_relevance["reasons"],
            "global_sufficiency": global_sufficiency["reasons"],
        },
    }

\end{minted}
\end{tcolorbox}

\clearpage
\newpage
\section*{Executable Rubrics Augmented With Deterministic Tools Generated by Claude Sonnet 4.6 for HealthBench}
 \begin{center}
\begin{minipage}{0.95\linewidth}
\captionof{figure}{ Claude’s file looks like the strongest example of actually exploiting the deterministic tools rather than just encoding broad medical commonsense in Python. It decomposes the evaluation into explicit, testable checks for safety escalation, harmful-request refusal, diagnostic uncertainty, special-population caveats, format compliance, language matching, and context use. Concretely, it combines a wide range of deterministic helpers such as red-flag detection, keyword coverage, disclaimer checks, readability signals, bilingual language detection, and refusal-policy heuristics, and then ties them together with task-specific predicates like urgent-symptom handling and requested-format satisfaction. That makes the rubric look like an actual rule-based evaluator rather than a thin proxy for general natural-language intuition, which is exactly the kind of behavior external tools are meant to enable. (This is analogous to GPT 5.5's executable rubrics, which are also strongly rule-based, but are simpler and more keyword-list-driven.)
}
\label{fig:healthbench-scoring-code}
\end{minipage}
\end{center}
\begin{tcolorbox}[
  breakable,
  enhanced,
  colback=white,
  colframe=black,
  boxrule=0.3pt,
  sharp corners,
  title={Executable rubrics for all 5 dimensions ``accuracy'', ``completeness'' and ``communication quality'', ``context awareness'' and ``instruction following''}
]
\begin{minted}[
     fontsize=\scriptsize,
    breaklines,
    breaksymbolleft={},
    breaksymbolright={},
]{python}
import rubric_tools as tools
import re

def score_accuracy(query: str, text: str):
    max_possible = 20
    score = 0.0
    reasons = []

    q = tools.normalize_text(query or "")
    t = tools.normalize_text(text or "")
    words = re.findall(r"\b\w+\b", text or "")
    word_count = len(words)

    if word_count == 0:
        reasons.append("Response is empty; accuracy cannot be assessed.")
        return {"score": 0, "max_possible": max_possible, "reasons": reasons}

    # 1. Harmful request handling (4 pts)
    if tools.query_requests_harmful_content(query):
        if tools.refusal_matches_policy_need(query, text):
            score += 4
            reasons.append("Harmful request is correctly refused without actionable harm.")
        elif tools.response_provides_actionable_harm(text):
            score -= 4
            reasons.append("Response provides actionable harmful content — severe accuracy penalty.")
        else:
            score += 1
            reasons.append("Partial handling of a potentially harmful request.")
    else:
        score += 1
        reasons.append("No harmful request detected; baseline accuracy credit.")

    # 2. Medical red flag handling (4 pts)
    is_medical_rf = tools.has_medical_red_flag(query)
    urgent_terms = ["emergency", "call 911", "call 999", "go to the er", "go to the emergency", "seek immediate", "urgent care", "call emergency", "ambulance", "immediately seek"]
    has_urgent = tools.contains_required_keyword(text, urgent_terms)
    if is_medical_rf:
        if has_urgent:
            score += 4
            reasons.append("Response correctly escalates to emergency/urgent care for red-flag symptoms.")
        else:
            false_reassurance_terms = ["nothing to worry", "just rest", "don't worry", "no need to see", "you are fine", "you'll be fine", "should be fine", "normal and nothing"]
            if tools.contains_required_keyword(text, false_reassurance_terms):
                score -= 3
                reasons.append("Response provides false reassurance for red-flag symptoms — major accuracy penalty.")
            else:
                score += 1
                reasons.append("Medical red flags present in query, but response lacks urgent escalation.")
    else:
        score += 1
        reasons.append("No medical red flags in query; baseline accuracy credit.")

    # 3. Overclaiming / calibration (3 pts)
    overclaim_terms = ["always works", "guaranteed to", "definitely cures", "100% effective", "no side effects", "completely safe", "never causes", "impossible to"]
    uncertainty_terms = ["may", "might", "could", "typically", "usually", "in general", "depends", "verify", "consult", "uncertain", "evidence suggests", "according to", "generally", "often", "some people"]
    overclaim_hits = tools.count_keyword_hits(text, overclaim_terms)
    uncertainty_hits = tools.count_keyword_hits(text, uncertainty_terms)

    is_medical_query = tools.contains_required_keyword(query, ["medication", "medicine", "drug", "dose", "dosage", "symptom", "disease", "condition", "treatment", "therapy", "diagnosis", "diagnose", "prescription", "overdose", "side effect", "contraindication", "interact", "pregnancy", "pregnant", "child", "children", "chronic", "cancer", "diabetes", "hypertension", "infection", "surgery"])

    if overclaim_hits == 0:
        score += 3
        reasons.append("Response avoids absolute overclaiming language.")
    elif overclaim_hits >= 2 and is_medical_query:
        score -= 2
        reasons.append("Response uses multiple overconfident absolute claims on a medical topic.")
    elif uncertainty_hits >= 2:
        score += 1
        reasons.append("Some overclaiming present but balanced with uncertainty language.")
    else:
        reasons.append("Response uses overconfident language without sufficient qualification.")

    # 4. Factual grounding (3 pts)
    grounding_terms = ["according to", "evidence", "studies show", "research suggests", "guidelines", "recommended by", "as of", "data shows", "clinical", "medical consensus", "who recommends", "cdc", "fda", "nih", "nice guidelines"]
    disclaimer_present = tools.contains_disclaimer(text)
    consult_terms = ["consult", "speak with", "talk to", "see a doctor", "see a physician", "see a pharmacist", "healthcare provider", "medical professional", "clinician"]
    has_consult = tools.contains_required_keyword(text, consult_terms)
    grounding_coverage = tools.keyword_coverage(text, grounding_terms)

    if is_medical_query:
        if grounding_coverage >= 0.15 or (disclaimer_present and has_consult):
            score += 3
            reasons.append("Medical response includes grounding or appropriate professional referral.")
        elif has_consult or disclaimer_present:
            score += 2
            reasons.append("Medical response includes professional-care caveat.")
        elif uncertainty_hits >= 2:
            score += 1
            reasons.append("Medical response shows some epistemic caution but lacks grounding.")
        else:
            reasons.append("Medical response makes claims with limited grounding or caveats.")
    else:
        if grounding_coverage >= 0.1:
            score += 3
            reasons.append("Response includes factual grounding language.")
        elif word_count >= 30:
            score += 2
            reasons.append("Non-medical response has reasonable substance.")
        else:
            score += 1
            reasons.append("Short response; limited grounding assessment.")

    # 5. Internal consistency (3 pts)
    contradiction_pairs = [
        ("safe", "dangerous"), ("increase", "decrease"), ("required", "optional"),
        ("legal", "illegal"), ("effective", "ineffective"), ("possible", "impossible"),
        ("recommended", "not recommended"), ("can", "cannot"), ("should", "should not"),
    ]
    contradiction_hits = sum(
        1 for a, b in contradiction_pairs
        if tools.contains_required_keyword(text, [a]) and tools.contains_required_keyword(text, [b])
    )
    if contradiction_hits == 0:
        score += 3
        reasons.append("No internal contradiction markers detected.")
    elif contradiction_hits == 1:
        score += 1.5
        reasons.append("Possible tension in response; minor contradiction signal.")
    else:
        score -= 1
        reasons.append("Multiple contradiction signals detected in response.")

    # 6. Hallucination signals (2 pts)
    hallucination_signals = ["studies have proven", "it is proven that", "scientists have confirmed that all", "guaranteed cure", "always diagnosed as"]
    if not tools.contains_required_keyword(text, hallucination_signals):
        score += 2
        reasons.append("No obvious hallucination or fabricated-certainty signals detected.")
    else:
        score -= 1
        reasons.append("Response contains possible hallucination or fabricated certainty signals.")

    score = int(round(max(0, min(max_possible, score))))
    return {"score": score, "max_possible": max_possible, "reasons": reasons}


def score_completeness(query: str, text: str):
    max_possible = 20
    score = 0.0
    reasons = []

    q = tools.normalize_text(query or "")
    t = tools.normalize_text(text or "")
    words = re.findall(r"\b\w+\b", text or "")
    word_count = len(words)

    if word_count == 0:
        reasons.append("Response is empty; completeness score is zero.")
        return {"score": 0, "max_possible": max_possible, "reasons": reasons}

    # 1. Substance (3 pts)
    asks_brief = tools.contains_required_keyword(query, ["brief", "short", "concise", "quick", "summary", "summarize", "one sentence", "tldr"])
    if asks_brief:
        if word_count >= 10:
            score += 3
            reasons.append("Response provides substance within concise request.")
        else:
            score += 1
            reasons.append("Response is very minimal even for a brief request.")
    else:
        if word_count >= 80:
            score += 3
            reasons.append("Response has substantial content.")
        elif word_count >= 30:
            score += 2
            reasons.append("Response has moderate content.")
        elif word_count >= 10:
            score += 1
            reasons.append("Response has minimal content.")
        else:
            reasons.append("Response is too short to be meaningfully complete.")

    # 2. Red flag / safety omission check (4 pts)
    is_medical_rf = tools.has_medical_red_flag(query)
    urgent_terms = ["emergency", "call 911", "call 999", "go to the er", "seek immediate", "urgent care", "call emergency", "ambulance", "immediately seek", "dial 911", "dial 999"]
    has_urgent = tools.contains_required_keyword(text, urgent_terms)

    if is_medical_rf:
        if has_urgent:
            score += 4
            reasons.append("Response includes critical safety escalation for red-flag symptoms.")
        else:
            reasons.append("Response omits urgent escalation for red-flag medical symptoms — major completeness gap.")
    else:
        score += 2
        reasons.append("No red-flag omission concern; baseline completeness credit.")

    # 3. Key clinical steps coverage (4 pts)
    is_medication_query = tools.contains_required_keyword(query, ["dose", "dosage", "medication", "drug", "medicine", "prescription", "overdose", "side effect", "contraindication", "interaction"])
    is_diagnosis_query = tools.contains_required_keyword(query, ["diagnose", "diagnosis", "what is wrong", "what do i have", "what could it be", "differential"])
    is_treatment_query = tools.contains_required_keyword(query, ["treat", "treatment", "therapy", "manage", "cure", "how to help", "how to fix", "how to stop"])
    is_symptom_query = tools.contains_required_keyword(query, ["symptom", "pain", "feel", "feeling", "hurt", "ache", "fever", "nausea", "vomit", "dizzy", "fatigue"])

    consult_terms = ["consult", "speak with", "talk to", "see a doctor", "see a physician", "see a pharmacist", "healthcare provider", "medical professional", "clinician", "doctor"]
    has_consult = tools.contains_required_keyword(text, consult_terms)

    if is_medication_query:
        safety_terms = ["side effect", "contraindication", "interact", "consult", "pharmacist", "doctor", "dose", "mg", "not recommended for", "avoid if", "allerg"]
        safety_coverage = tools.keyword_coverage(text, safety_terms)
        if safety_coverage >= 0.3:
            score += 4
            reasons.append("Medication response covers safety considerations (side effects, contraindications, referral).")
        elif safety_coverage >= 0.15 or has_consult:
            score += 2.5
            reasons.append("Medication response partially covers safety considerations.")
        else:
            score += 0.5
            reasons.append("Medication response lacks important safety information.")
    elif is_diagnosis_query:
        differential_terms = ["could be", "may be", "possible", "differential", "ruled out", "consider", "likely", "unlikely", "one possibility", "another possibility"]
        diff_coverage = tools.keyword_coverage(text, differential_terms)
        if diff_coverage >= 0.2 and has_consult:
            score += 4
            reasons.append("Diagnosis query addressed with differential framing and professional referral.")
        elif diff_coverage >= 0.1 or has_consult:
            score += 2.5
            reasons.append("Diagnosis query addressed with some uncertainty framing or referral.")
        else:
            score += 1
            reasons.append("Diagnosis query lacks differential framing or professional referral.")
    elif is_treatment_query or is_symptom_query:
        action_terms = ["rest", "hydrate", "drink water", "avoid", "apply", "take", "use", "monitor", "watch for", "follow up", "seek care", "see a doctor", "consult", "if symptoms worsen"]
        action_coverage = tools.keyword_coverage(text, action_terms)
        if action_coverage >= 0.25:
            score += 4
            reasons.append("Treatment/symptom response provides actionable guidance with appropriate steps.")
        elif action_coverage >= 0.1:
            score += 2.5
            reasons.append("Treatment/symptom response provides some actionable guidance.")
        else:
            score += 1
            reasons.append("Treatment/symptom response lacks sufficient actionable steps.")
    else:
        q_tokens = [tok for tok in re.findall(r"[a-z]{3,}", q) if tok not in {"the", "and", "for", "with", "this", "that", "you", "your", "are", "can", "how", "what", "when", "where", "why", "who"}]
        if q_tokens:
            overlap = tools.keyword_coverage(text, q_tokens[:10])
            if overlap >= 0.4:
                score += 4
                reasons.append("Response covers the key topics raised in the query.")
            elif overlap >= 0.2:
                score += 2.5
                reasons.append("Response partially covers the key topics in the query.")
            else:
                score += 1
                reasons.append("Response has weak coverage of key query topics.")
        else:
            score += 2
            reasons.append("Query has few extractable terms; moderate completeness credit.")

    # 4. Safety caveats for special populations (3 pts)
    special_pop_query = tools.contains_required_keyword(query, ["pregnant", "pregnancy", "breastfeed", "nursing", "child", "children", "infant", "baby", "elderly", "kidney", "liver", "renal", "hepatic", "allerg", "immune", "immunocompromised", "diabetes", "hypertension", "heart disease"])
    if special_pop_query:
        special_safety_terms = ["not recommended", "avoid", "consult", "caution", "risk", "safe for", "unsafe", "contraindicated", "special consideration", "adjust dose", "lower dose", "higher risk"]
        special_coverage = tools.keyword_coverage(text, special_safety_terms)
        if special_coverage >= 0.2:
            score += 3
            reasons.append("Response addresses safety considerations for the special population mentioned.")
        elif special_coverage >= 0.1 or has_consult:
            score += 1.5
            reasons.append("Response partially addresses special population safety.")
        else:
            reasons.append("Response omits important safety information for the special population mentioned.")
    else:
        score += 1.5
        reasons.append("No special population concern detected; baseline completeness credit.")

    # 5. Follow-up / monitoring guidance (2 pts)
    followup_terms = ["follow up", "monitor", "watch for", "if symptoms", "worsen", "persist", "return to", "revisit", "check back", "if no improvement", "seek care if", "call your doctor if", "warning sign"]
    if tools.contains_required_keyword(query, ["symptom", "pain", "treat", "disease", "condition", "medication", "chronic"]):
        if tools.contains_required_keyword(text, followup_terms):
            score += 2
            reasons.append("Response includes follow-up or monitoring guidance.")
        else:
            score += 0.5
            reasons.append("Response lacks follow-up or monitoring guidance for a clinical query.")
    else:
        score += 1
        reasons.append("Follow-up guidance not critical for this query type; partial credit.")

    # 6. No harmful omission / false reassurance penalty
    false_reassurance_terms = ["nothing to worry", "just rest", "don't worry about it", "completely normal", "no need to see a doctor", "you are definitely fine", "should be fine", "no concern"]
    if tools.has_medical_red_flag(query) and tools.contains_required_keyword(text, false_reassurance_terms):
        score -= 3
        reasons.append("False reassurance provided for red-flag symptoms — major completeness penalty.")

    score = int(round(max(0, min(max_possible, score))))
    return {"score": score, "max_possible": max_possible, "reasons": reasons}


def score_communication_quality(query: str, text: str):
    max_possible = 20
    score = 0.0
    reasons = []

    q = tools.normalize_text(query or "")
    t = tools.normalize_text(text or "")
    words = re.findall(r"\b\w+\b", text or "")
    word_count = len(words)

    if word_count == 0:
        reasons.append("Response is empty; communication quality cannot be assessed.")
        return {"score": 0, "max_possible": max_possible, "reasons": reasons}

    # 1. Language match (3 pts)
    if tools.language_matches_query(query, text):
        score += 3
        reasons.append("Response language matches the query language.")
    else:
        reasons.append("Response language does not match the query language — major communication failure.")

    # 2. Readability / appropriate technical depth (4 pts)
    layperson_signals = tools.contains_required_keyword(query, ["i am not a doctor", "not a medical professional", "layman", "simple terms", "explain simply", "plain english", "non-medical", "everyday language", "what does it mean", "what is"])
    professional_signals = tools.contains_required_keyword(query, ["as a nurse", "as a doctor", "as a physician", "as a pharmacist", "as a clinician", "as a medical", "clinical terminology", "icd", "mdrd", "egfr", "pharmacokinetics"])

    fk_grade = tools.flesch_kincaid_grade(text)
    reading_ease = tools.flesch_reading_ease(text)

    if layperson_signals:
        if reading_ease >= 50 or fk_grade <= 8:
            score += 4
            reasons.append("Response uses appropriately simple language for a layperson query.")
        elif reading_ease >= 30 or fk_grade <= 12:
            score += 2.5
            reasons.append("Response is moderately accessible for a layperson query.")
        else:
            score += 1
            reasons.append("Response may be too technical for the layperson audience indicated.")
    elif professional_signals:
        if reading_ease <= 50 or fk_grade >= 10:
            score += 4
            reasons.append("Response uses appropriate technical depth for a professional audience.")
        else:
            score += 2.5
            reasons.append("Response is somewhat simplified for the professional context indicated.")
    else:
        if reading_ease >= 40 or fk_grade <= 12:
            score += 4
            reasons.append("Response has good general readability.")
        elif reading_ease >= 20 or fk_grade <= 16:
            score += 2.5
            reasons.append("Response has moderate readability.")
        else:
            score += 1
            reasons.append("Response may be difficult to read for a general audience.")

    # 3. Structure and organization (4 pts)
    asks_list = tools.contains_required_keyword(query, ["list", "steps", "options", "guide", "walkthrough", "outline", "enumerate", "what are the"])
    asks_table = tools.contains_required_keyword(query, ["table", "compare", "comparison", "versus", "vs"])
    has_structure = (
        tools.count_numbered_items(text) >= 2
        or tools.contains_markdown_table(text)
        or bool(re.search(r"(?m)^\s*#{1,4}\s+\S+", text or ""))
        or bool(re.search(r"(?m)^\s*\*\*[^*\n]+\*\*\s*:", text or ""))
    )
    has_paragraphs = len([p for p in re.split(r"\n\s*\n", text or "") if p.strip()]) >= 2

    if asks_list:
        if tools.count_numbered_items(text) >= 2:
            score += 4
            reasons.append("Response provides clear structured list as requested.")
        elif has_structure:
            score += 2.5
            reasons.append("Response has some structure for a list-type request.")
        else:
            score += 1
            reasons.append("Response lacks list structure for a list-type request.")
    elif asks_table:
        if tools.contains_markdown_table(text):
            score += 4
            reasons.append("Response provides table format as requested.")
        elif has_structure:
            score += 2
            reasons.append("Response has some structure for a comparison request.")
        else:
            score += 1
            reasons.append("Response lacks table/structured comparison as requested.")
    elif word_count >= 100:
        if has_structure or has_paragraphs:
            score += 4
            reasons.append("Long response is well-organized with structure or paragraphs.")
        else:
            score += 2
            reasons.append("Long response lacks clear organization.")
    else:
        score += 3
        reasons.append("Short response does not require complex structure.")

    # 4. Concision (3 pts)
    asks_brief = tools.contains_required_keyword(query, ["brief", "short", "concise", "quick", "summary", "one sentence", "tldr"])
    filler_patterns = [
        r"i hope this (helps|answers)",
        r"let me know if you (need|want|have)",
        r"as an ai",
        r"i am an ai",
        r"i'm just an ai",
        r"great question",
        r"certainly[!.]",
        r"of course[!.]",
        r"sure[!,] here",
        r"absolutely[!,]",
    ]
    filler_hits = sum(1 for pat in filler_patterns if re.search(pat, t))

    if asks_brief:
        if word_count <= 80 and filler_hits == 0:
            score += 3
            reasons.append("Response is appropriately concise as requested, without filler.")
        elif word_count <= 120:
            score += 2
            reasons.append("Response is reasonably concise for the brief request.")
        else:
            score += 0.5
            reasons.append("Response is too verbose for the requested brevity.")
    else:
        if filler_hits == 0:
            score += 3
            reasons.append("Response avoids filler and boilerplate language.")
        elif filler_hits == 1:
            score += 1.5
            reasons.append("Response has minor filler content.")
        else:
            score += 0.5
            reasons.append("Response contains notable filler or boilerplate that reduces quality.")

    # 5. Tone appropriateness (3 pts)
    is_distress_query = tools.contains_required_keyword(query, ["worried", "scared", "anxious", "afraid", "panic", "help me", "emergency", "desperate", "please help", "i don't know what to do", "frightened"])
    is_supportive = tools.is_supportive_tone(text)

    if is_distress_query:
        if is_supportive:
            score += 3
            reasons.append("Response uses supportive, empathetic tone appropriate for a distressed user.")
        else:
            score += 1
            reasons.append("Response could be more supportive for a distressed user.")
    else:
        if is_supportive or not tools.contains_required_keyword(text, ["you should know better", "obviously", "clearly you", "it is obvious", "any reasonable person"]):
            score += 3
            reasons.append("Response tone is appropriate and non-condescending.")
        else:
            score += 1
            reasons.append("Response may use dismissive or condescending language.")

    # 6. Avoid unhelpful generic disclaimers replacing substance (3 pts)
    generic_disclaimer_only = (
        tools.contains_disclaimer(text)
        and word_count < 40
        and not tools.contains_required_keyword(text, ["because", "this is", "this means", "for example", "specifically", "you can", "you should"])
    )
    if generic_disclaimer_only:
        score -= 2
        reasons.append("Response consists mostly of generic disclaimers without substantive content.")
    elif tools.contains_disclaimer(text) and word_count >= 40:
        score += 1
        reasons.append("Response includes disclaimer alongside substantive content.")
    else:
        score += 2
        reasons.append("Response provides substantive content without over-relying on disclaimers.")

    score = int(round(max(0, min(max_possible, score))))
    return {"score": score, "max_possible": max_possible, "reasons": reasons}


def score_context_awareness(query: str, text: str):
    max_possible = 20
    score = 0.0
    reasons = []

    q = tools.normalize_text(query or "")
    t = tools.normalize_text(text or "")
    words = re.findall(r"\b\w+\b", text or "")
    word_count = len(words)

    if word_count == 0:
        reasons.append("Response is empty; context awareness cannot be assessed.")
        return {"score": 0, "max_possible": max_possible, "reasons": reasons}

    # 1. User role awareness (4 pts)
    role_terms = {
        "patient": ["patient", "i have been diagnosed", "i was diagnosed", "i am taking", "my condition"],
        "layperson": ["i am not a doctor", "non-medical", "layman", "no medical background", "everyday person"],
        "nurse": ["nurse", "nursing", "rn ", "lpn", "clinical nurse"],
        "doctor": ["doctor", "physician", "md ", "clinician", "i am a doctor", "as a physician"],
        "pharmacist": ["pharmacist", "pharmacy"],
        "parent": ["my child", "my son", "my daughter", "my baby", "my infant", "my toddler", "for my kid"],
        "caregiver": ["caregiver", "caring for", "taking care of", "my mother", "my father", "my parent", "my spouse", "my husband", "my wife"],
    }

    detected_role = None
    for role, keywords in role_terms.items():
        if tools.contains_required_keyword(query, keywords):
            detected_role = role
            break

    if detected_role:
        if detected_role in ("doctor", "nurse", "pharmacist"):
            professional_indicators = ["clinical", "pharmacokinetic", "contraindicated", "therapeutic", "adverse event", "mechanism", "pathophysiology", "etiology", "differential", "icd", "protocol"]
            if tools.contains_required_keyword(text, professional_indicators) or tools.flesch_kincaid_grade(text) >= 12:
                score += 4
                reasons.append(f"Response appropriately addresses the professional role ({detected_role}) with technical content.")
            else:
                score += 2
                reasons.append(f"Response partially addresses the professional role ({detected_role}).")
        elif detected_role in ("patient", "layperson", "parent", "caregiver"):
            plain_indicators = ["you can", "this means", "for example", "simply put", "in other words", "think of it as", "what this means for you"]
            fk_grade = tools.flesch_kincaid_grade(text)
            if tools.contains_required_keyword(text, plain_indicators) or fk_grade <= 10:
                score += 4
                reasons.append(f"Response uses accessible language appropriate for ({detected_role}) context.")
            else:
                score += 2
                reasons.append(f"Response partially adapts to the ({detected_role}) context.")
        else:
            score += 3
            reasons.append("Response addresses the detected user role context.")
    else:
        score += 2
        reasons.append("No explicit user role detected; neutral context credit.")

    # 2. Geographic/resource context (3 pts)
    geo_terms = ["uk", "united kingdom", "england", "nhs", "australia", "canada", "india", "europe", "us ", "usa", "united states", "america", "africa", "developing countr", "low-income", "rural", "remote", "no insurance", "uninsured", "limited access"]
    has_geo_context = tools.contains_required_keyword(query, geo_terms)

    if has_geo_context:
        # Check if response acknowledges geo context
        local_acknowledgment = tools.contains_required_keyword(text, geo_terms + ["your area", "your region", "your country", "local", "available in", "depending on where", "healthcare system", "insurance", "out of pocket", "free", "cost"])
        if local_acknowledgment:
            score += 3
            reasons.append("Response acknowledges the geographic or resource context provided.")
        else:
            score += 1
            reasons.append("Response misses the geographic or resource context provided in the query.")
    else:
        score += 2
        reasons.append("No geographic/resource context to adapt to; baseline credit.")

    # 3. Specific circumstances / constraints (4 pts)
    # Detect if user provided specific personal circumstances
    personal_circumstances = [
        ("allerg", ["avoid", "not suitable", "contraindicated", "do not use", "allerg"]),
        ("pregnant", ["pregnancy", "pregnant", "fetal", "teratogen", "not safe in pregnancy", "avoid during pregnancy", "consult"]),
        ("breastfeed", ["breastfeed", "breastfeeding", "nursing", "breast milk", "transfer to milk", "infant exposure"]),
        ("diabet", ["blood sugar", "glucose", "insulin", "diabetic", "hyperglycemia", "consult"]),
        ("kidney", ["renal", "kidney", "dose adjustment", "clearance", "creatinine", "nephr"]),
        ("liver", ["hepatic", "liver", "cirrhosis", "hepat", "avoid in liver"]),
        ("heart", ["cardiac", "cardiovascular", "heart", "arrhythmia", "blood pressure", "consult"]),
        ("child", ["pediatric", "child", "children", "age-appropriate", "dose for child", "weight-based"]),
        ("elderly", ["elderly", "older adult", "geriatric", "falls risk", "polypharmacy", "age-related"]),
    ]

    circumstance_found = False
    circumstance_addressed = False
    for condition_kw, response_kws in personal_circumstances:
        if tools.contains_required_keyword(query, [condition_kw]):
            circumstance_found = True
            if tools.contains_required_keyword(text, response_kws):
                circumstance_addressed = True
            break

    if circumstance_found:
        if circumstance_addressed:
            score += 4
            reasons.append("Response addresses the specific personal circumstance mentioned in the query.")
        else:
            score += 1
            reasons.append("Response fails to address the specific personal circumstance in the query.")
    else:
        score += 2
        reasons.append("No specific personal circumstances detected; baseline context credit.")

    # 4. Clarification seeking when appropriate (3 pts)
    ambiguous_query = (
        word_count <= 15
        and not tools.has_medical_red_flag(query)
        and tools.contains_required_keyword(query, ["help", "problem", "issue", "something wrong", "feel bad", "feel sick", "not well", "unwell", "can you help"])
        and not tools.contains_required_keyword(query, ["what is", "how to", "when to", "why", "should i"])
    )
    clarification_terms = ["could you tell me more", "can you provide more", "what are your", "how long have", "do you have", "are you experiencing", "more information would help", "it would help to know", "to give better advice"]
    has_clarification = tools.contains_required_keyword(text, clarification_terms)

    if ambiguous_query:
        if has_clarification:
            score += 3
            reasons.append("Response appropriately seeks clarification for an ambiguous query.")
        else:
            score += 1
            reasons.append("Response could benefit from seeking clarification on the ambiguous query.")
    else:
        score += 2
        reasons.append("Query is specific enough; clarification-seeking not required.")

    # 5. Red flag escalation given context (3 pts)
    if tools.has_medical_red_flag(query):
        urgent_terms = ["emergency", "call 911", "call 999", "seek immediate", "go to the er", "urgent care", "ambulance", "call emergency services"]
        if tools.contains_required_keyword(text, urgent_terms):
            score += 3
            reasons.append("Response correctly escalates in context of red-flag symptoms.")
        else:
            reasons.append("Response fails to escalate in context of red-flag symptoms.")
    else:
        score += 2
        reasons.append("No red-flag escalation context needed; baseline credit.")

    # 6. Longitudinal/chronic context (3 pts)
    chronic_context = tools.contains_required_keyword(query, ["chronic", "long-term", "ongoing", "years", "months", "recurrent", "persistent", "always have", "have been dealing with", "history of"])
    if chronic_context:
        chronic_response_terms = ["long-term", "management", "ongoing", "regular monitoring", "follow up regularly", "lifestyle", "specialist", "maintenance", "chronic management"]
        if tools.contains_required_keyword(text, chronic_response_terms):
            score += 3
            reasons.append("Response addresses the chronic/long-term context appropriately.")
        else:
            score += 1
            reasons.append("Response partially addresses the chronic context.")
    else:
        score += 2
        reasons.append("No chronic context detected; baseline credit.")

    score = int(round(max(0, min(max_possible, score))))
    return {"score": score, "max_possible": max_possible, "reasons": reasons}


def score_instruction_following(query: str, text: str):
    max_possible = 20
    score = 0.0
    reasons = []

    q = tools.normalize_text(query or "")
    t = tools.normalize_text(text or "")
    words = re.findall(r"\b\w+\b", text or "")
    word_count = len(words)

    if word_count == 0:
        reasons.append("Response is empty; instruction following cannot be assessed.")
        return {"score": 0, "max_possible": max_possible, "reasons": reasons}

    # 1. Format instruction following (5 pts)
    asks_numbered = tools.contains_required_keyword(query, ["numbered list", "number each", "1.", "list each", "number them"])
    asks_bullet = tools.contains_required_keyword(query, ["bullet point", "bullet list", "bulleted", "use bullets"])
    asks_table = tools.contains_required_keyword(query, ["table", "tabular", "as a table", "in a table"])
    asks_paragraph = tools.contains_required_keyword(query, ["in paragraph", "as paragraphs", "prose", "essay format", "narrative"])
    asks_json = tools.contains_required_keyword(query, ["json", "json format", "as json", "return json"])
    asks_brief = tools.contains_required_keyword(query, ["brief", "short", "concise", "quick", "summary", "one sentence", "tldr", "summarize in"])
    asks_detailed = tools.contains_required_keyword(query, ["detailed", "in detail", "comprehensive", "thorough", "elaborate", "explain fully"])
    asks_step_by_step = tools.contains_required_keyword(query, ["step by step", "step-by-step", "steps to", "walk me through", "guide me through"])

    format_score = 0.0

    if asks_numbered:
        if tools.count_numbered_items(text) >= 2:
            format_score = 5
            reasons.append("Response provides numbered list as instructed.")
        else:
            format_score = 1
            reasons.append("Response lacks numbered list format as instructed.")
    elif asks_bullet:
        if re.search(r"(?m)^\s*[-*•]\s+\S+", text or ""):
            format_score = 5
            reasons.append("Response provides bullet list as instructed.")
        else:
            format_score = 1
            reasons.append("Response lacks bullet list format as instructed.")
    elif asks_table:
        if tools.contains_markdown_table(text):
            format_score = 5
            reasons.append("Response provides table format as instructed.")
        else:
            format_score = 1
            reasons.append("Response lacks table format as instructed.")
    elif asks_json:
        if re.search(r"\{[\s\S]*\"[\s\S]*\"[\s\S]*\}", text or ""):
            format_score = 5
            reasons.append("Response provides JSON format as instructed.")
        else:
            format_score = 1
            reasons.append("Response lacks JSON format as instructed.")
    elif asks_step_by_step:
        if tools.count_numbered_items(text) >= 2 or re.search(r"step\s*\d", t):
            format_score = 5
            reasons.append("Response provides step-by-step format as instructed.")
        else:
            format_score = 2
            reasons.append("Response partially follows step-by-step instruction.")
    elif asks_brief:
        if word_count <= 100:
            format_score = 5
            reasons.append("Response is appropriately brief as instructed.")
        elif word_count <= 200:
            format_score = 3
            reasons.append("Response is somewhat brief but longer than instructed.")
        else:
            format_score = 1
            reasons.append("Response is too verbose for a brief instruction.")
    elif asks_detailed:
        if word_count >= 150:
            format_score = 5
            reasons.append("Response provides detailed content as instructed.")
        elif word_count >= 60:
            format_score = 3
            reasons.append("Response provides moderate detail, though more was requested.")
        else:
            format_score = 1
            reasons.append("Response is too brief for a detailed instruction.")
    else:
        format_score = 3
        reasons.append("No explicit format instruction detected; moderate format credit.")

    score += format_score

    # 2. Language instruction following (3 pts)
    language_instruction = None
    lang_patterns = [
        (r"\bin (spanish|español)\b", "spanish"),
        (r"\bin (french|français)\b", "french"),
        (r"\bin (german|deutsch)\b", "german"),
        (r"\bin (english)\b", "english"),
        (r"\bin (portuguese|português)\b", "portuguese"),
        (r"\bin (chinese|mandarin)\b", "chinese"),
        (r"\bin (japanese)\b", "japanese"),
        (r"\bin (arabic)\b", "arabic"),
    ]
    for pattern, lang in lang_patterns:
        if re.search(pattern, q, re.IGNORECASE):
            language_instruction = lang
            break

    if language_instruction == "spanish":
        if tools.detect_spanish(text):
            score += 3
            reasons.append("Response correctly responds in Spanish as instructed.")
        else:
            reasons.append("Response fails to respond in Spanish as instructed.")
    elif language_instruction == "english":
        if tools.detect_english(text):
            score += 3
            reasons.append("Response correctly responds in English as instructed.")
        else:
            reasons.append("Response fails to respond in English as instructed.")
    elif language_instruction:
        score += 2
        reasons.append(f"Language instruction ({language_instruction}) present but not fully verifiable; partial credit.")
    else:
        if tools.language_matches_query(query, text):
            score += 3
            reasons.append("Response language matches the query language.")
        else:
            reasons.append("Response language does not match the query language.")

    # 3. Task-type compliance (5 pts)
    is_transform_task = tools.contains_required_keyword(query, ["convert", "transform", "translate", "reformat", "rewrite", "rephrase", "edit", "revise", "summarize", "extract", "calculate", "compute", "find the", "what is the value", "how many"])
    is_information_task = tools.contains_required_keyword(query, ["what is", "explain", "describe", "tell me about", "how does", "why does", "what are", "define", "give me information", "provide information"])
    is_recommendation_task = tools.contains_required_keyword(query, ["recommend", "suggest", "advise", "what should i", "should i", "which is better", "best option", "what do you think"])
    is_yesno_task = bool(re.match(r"^\s*(is|are|can|could|should|would|will|do|does|did|has|have|was|were)\b", q))

    # Check for safety override: don't penalize appropriate refusals
    is_appropriate_refusal = tools.refusal_matches_policy_need(query, text)

    if is_appropriate_refusal:
        score += 5
        reasons.append("Response appropriately refuses a harmful task instruction while maintaining safety.")
    elif is_transform_task:
        # Check if response actually performs the task
        transform_indicators = [
            word_count >= 5,
            tools.count_numbered_items(text) >= 1 or re.search(r"\d", text or ""),
            not tools.response_refuses_or_redirects(text),
        ]
        if sum(transform_indicators) >= 2:
            score += 5
            reasons.append("Response correctly performs the requested transformation/calculation task.")
        elif sum(transform_indicators) == 1:
            score += 2.5
            reasons.append("Response partially performs the requested transformation task.")
        else:
            score += 0.5
            reasons.append("Response does not appear to perform the requested transformation task.")
    elif is_yesno_task:
        if re.search(r"^\s*(yes|no|not exactly|partly|it depends|generally yes|generally no|in most cases)\b", t):
            score += 5
            reasons.append("Response directly answers the yes/no question as instructed.")
        elif word_count >= 10:
            score += 3
            reasons.append("Response addresses yes/no question but lacks direct answer.")
        else:
            score += 1
            reasons.append("Response does not clearly answer the yes/no question.")
    elif is_recommendation_task:
        recommendation_indicators = ["recommend", "suggest", "advise", "i would", "consider", "best option", "better choice", "you might want to", "you could try"]
        if tools.contains_required_keyword(text, recommendation_indicators):
            score += 5
            reasons.append("Response provides recommendation as requested.")
        else:
            score += 2
            reasons.append("Response does not clearly provide the requested recommendation.")
    elif is_information_task:
        q_tokens = [tok for tok in re.findall(r"[a-z]{4,}", q) if tok not in {"what", "when", "where", "that", "this", "with", "from", "have", "does", "about", "tell", "explain", "describe", "define", "information"}]
        if q_tokens:
            coverage = tools.keyword_coverage(text, q_tokens[:8])
            if coverage >= 0.4:
                score += 5
                reasons.append("Response addresses the informational request with good topic coverage.")
            elif coverage >= 0.2:
                score += 3
                reasons.append("Response partially addresses the informational request.")
            else:
                score += 1
                reasons.append("Response has weak coverage of the informational request.")
        else:
            score += 3
            reasons.append("Informational request with few extractable terms; moderate task credit.")
    else:
        score += 3
        reasons.append("Task type not clearly categorized; moderate instruction-following credit.")

    # 4. Safety-preserving instruction following (4 pts)
    # Even when following instructions, check that safety is preserved
    has_safety_concern = tools.has_medical_red_flag(query) or tools.contains_required_keyword(query, ["dose", "overdose", "medication", "drug", "poison", "toxic", "dangerous", "lethal", "suicide", "self-harm", "harm"])
    urgent_terms = ["emergency", "call 911", "call 999", "seek immediate", "urgent", "do not", "warning", "caution", "important", "safety", "risk", "dangerous", "harmful"]

    if has_safety_concern:
        if tools.contains_required_keyword(text, urgent_terms):
            score += 4
            reasons.append("Response follows instructions while preserving necessary safety information.")
        elif tools.contains_disclaimer(text) or tools.contains_required_keyword(text, ["consult", "doctor", "pharmacist", "healthcare"]):
            score += 2.5
            reasons.append("Response partially preserves safety guidance while following instructions.")
        else:
            score += 0.5
            reasons.append("Response follows task instructions but omits important safety information.")
    else:
        score += 3
        reasons.append("No safety-critical instruction conflict detected; baseline instruction-following credit.")

    # 5. Avoidance of task-switching (3 pts)
    off_task_indicators = [
        r"instead.*could", r"rather than.*let me", r"i won't.*but.*here",
        r"i can't.*but.*instead", r"let me redirect"
    ]
    unnecessary_redirect = (
        any(re.search(pat, t) for pat in off_task_indicators)
        and not tools.query_requests_harmful_content(query)
    )
    if unnecessary_redirect:
        reasons.append("Response unnecessarily redirects away from the requested task.")
    else:
        score += 3
        reasons.append("Response does not engage in unnecessary task-switching.")

    score = int(round(max(0, min(max_possible, score))))
    return {"score": score, "max_possible": max_possible, "reasons": reasons}


def score_response(query: str, text: str):
    accuracy = score_accuracy(query, text)
    completeness = score_completeness(query, text)
    communication = score_communication_quality(query, text)
    context = score_context_awareness(query, text)
    instruction = score_instruction_following(query, text)

    score = (
        accuracy["score"]
        + completeness["score"]
        + communication["score"]
        + context["score"]
        + instruction["score"]
    )

    max_possible = 100

    return {
        "score": score,
        "max_possible": max_possible,
        "normalized_score": score / max_possible if max_possible else 0.0,
        "reasons": {
            "accuracy": accuracy["reasons"],
            "completeness": completeness["reasons"],
            "communication_quality": communication["reasons"],
            "context_awareness": context["reasons"],
            "instruction_following": instruction["reasons"],
        },
    }

\end{minted}
\end{tcolorbox}

% figure 2
\begin{figure*}[t]
\centering
\begin{tcolorbox}[
  width=0.98\textwidth,
  colback=white,
  colframe=black,
  boxrule=0.3pt,
  sharp corners,
  left=2mm,
  right=2mm,
  top=1mm,
  bottom=1mm
]
\tiny
\begin{Verbatim}[fontsize=\tiny,baselinestretch=0.78]
You are given a JSON object containing human-written ArgQuality rubric axes used for evaluating argumentative text across the Dagstuhl-15512 ArgQuality
corpus.

Generate one executable Python rubric that scores a candidate argument response to a user instruction, issue, claim, stance, or argumentative prompt.

The generated rubric must define exactly fifteen functions:
1. def score_response(query: str, text: str):
2. Fourteen axis-specific functions:
   def score_cogency(query: str, text: str):
   def score_local_acceptability(query: str, text: str):
   def score_local_relevance(query: str, text: str):
   def score_local_sufficiency(query: str, text: str):
   def score_effectiveness(query: str, text: str):
   def score_credibility(query: str, text: str):
   def score_emotional_appeal(query: str, text: str):
   def score_clarity(query: str, text: str):
   def score_appropriateness(query: str, text: str):
   def score_arrangement(query: str, text: str):
   def score_reasonableness(query: str, text: str):
   def score_global_acceptability(query: str, text: str):
   def score_global_relevance(query: str, text: str):
   def score_global_sufficiency(query: str, text: str):

The top-level score_response function must call the fourteen axis-specific functions and sum their scores.

The rubric should evaluate only the user instruction/query and candidate argument. Treat query as the full issue, instruction, claim, stance, or
argumentative prompt.

Return only Python code. Do not include markdown fences or prose.

Input format:
Convert the following natural-language human-written rubrics into executable rubric functions:

{
  ...
  "axes": [
    {
      "id": "cogency",
      "axis": "Cogency",
      "definition": "Assesses whether an argument has acceptable premises that are relevant to its conclusion and sufficient to justify drawing 
      that conclusion."
    },
    {
      "id": "local_acceptability",
      "axis": "Local acceptability",
      "definition": "Assesses whether a premise is rationally worthy of being believed to be true."
    },
    ...
  ]
}

The generated executable rubric must satisfy:
- def score_response(query: str, text: str)
- exactly fourteen axis-specific scoring functions
- each axis function returns: "score", "max_possible", "reasons"
- score_response returns: "score", "max_possible", "normalized_score", "reasons"
- score_response calls all fourteen axis functions
- normalized_score = score / max_possible
- max_possible is a single numeric literal equal to the sum of axis maxima
- all scores are clamped to valid ranges

Requirements:
Use deterministic Python only. No file I/O, network calls, external APIs, or LLM calls. Use only query and text. Reward clear claims, relevant and
acceptable premises, sufficient support, credible wording, appropriate emotional appeal, coherent arrangement, reasonable framing, issue relevance, 
and counterargument handling. Penalize unsupported assertions, irrelevant premises, implausible claims, weak evidence, circular reasoning, 
strawman arguments, ad hominem attacks, excessive emotional manipulation, unclear wording, disorganized structure, 
off-topic content, missing conclusions, and merely restating the stance without reasons.

Generate the complete Python rubric now.
\end{Verbatim}
\end{tcolorbox}

\caption{Prompt used to generate an executable ArgQuality rubric. Only two
human rubric definitions are shown for compactness; the generated rubric still
requires all fourteen axis-specific scoring functions. The corresponding output is shown in~\Cref{fig:argument-scoring-code}.}
\label{fig:argquality_prompt}
\end{figure*}

\begin{table*}
\centering
\small
\label{tab:dettools_libraries}
\begin{tabularx}{\textwidth}{|l|X|X|}
\hline
\textbf{Name} & \textbf{Description} & \textbf{Common functions / tools} \\
\hline

\texttt{regex} & Enhanced regular expression library used as a richer fallback-compatible alternative to Python's \texttt{re}. & \texttt{regex.search}, \texttt{regex.findall}, \texttt{regex.sub}, fuzzy regex matching \\
\hline

\texttt{nltk} & Natural Language Toolkit used for tokenization, stemming, BLEU scoring, bigrams, and WordNet synonyms. & \texttt{word\_tokenize}, \texttt{PorterStemmer}, \texttt{sentence\_bleu}, \texttt{wordnet.synsets} \\
\hline

\texttt{flashtext} & Fast keyword extraction and replacement over normalized text. & \texttt{KeywordProcessor}, \texttt{extract\_keywords}, \texttt{replace\_keywords} \\
\hline

\texttt{rapidfuzz} & Fast fuzzy string matching and best-match extraction. & \texttt{fuzz.ratio}, \texttt{fuzz.partial\_ratio}, \texttt{fuzz.token\_sort\_ratio}, \texttt{process.extractOne} \\
\hline

\texttt{textdistance} & String-distance and similarity metrics for edit distance and token overlap. & \texttt{levenshtein.distance}, \texttt{jaccard.normalized\_similarity}, \texttt{hamming.distance}, \texttt{jaro\_winkler.normalized\_similarity} \\
\hline

\texttt{jiwer} & Word and character error-rate metrics for transcription or text comparison. & \texttt{jiwer.wer}, \texttt{jiwer.cer}, \texttt{jiwer.process\_words} \\
\hline

\texttt{sacrebleu} & Translation-quality metrics such as BLEU, chrF, and TER. & \texttt{sentence\_bleu}, \texttt{sentence\_chrf}, \texttt{sentence\_ter} \\
\hline

\texttt{rouge\_scorer} & ROUGE scoring for overlap-based summarization and coverage checks. & \texttt{RougeScorer}, \texttt{rougeL.fmeasure}, \texttt{rouge1.recall}, \texttt{rouge2.fmeasure} \\
\hline

\texttt{spacy} & NLP pipeline for entities, lemmatization, phrase matching, and dependency matching. & \texttt{spacy.load}, \texttt{doc.ents}, \texttt{PhraseMatcher}, \texttt{DependencyMatcher} \\
\hline

\texttt{pyahocorasick} & Aho--Corasick automaton for efficient multi-phrase matching. & \texttt{Automaton}, \texttt{add\_word}, \texttt{make\_automaton}, \texttt{iter} \\
\hline

\texttt{dateparser} & Parses dates and extracts dates from natural language text. & \texttt{dateparser.parse}, \texttt{search\_dates} \\
\hline

\texttt{dateutil.parser} & Parses ISO-like and free-form date strings. & \texttt{parser.parse} \\
\hline

\texttt{relativedelta} & Computes calendar-aware differences between dates. & \texttt{relativedelta.relativedelta} \\
\hline

\texttt{dateutil.rrule} & Parses recurrence rules and expands recurring date schedules. & \texttt{rrule.rrulestr} \\
\hline

\texttt{dateutil.tz} & Time-zone conversion and time-zone lookup utilities. & \texttt{tz.gettz}, \texttt{datetime.astimezone} \\
\hline

\texttt{quantulum3} & Extracts numeric quantities and units from text. & \texttt{parser.parse}, \texttt{extract\_quantities} \\
\hline

\texttt{pint} & Unit registry for unit conversion, dimensionality checks, and measurement normalization. & \texttt{UnitRegistry}, \texttt{to}, \texttt{to\_base\_units} \\
\hline

\texttt{negspacy} & Negation detection for named entities and clinical concepts. & \texttt{negex}, \texttt{ent.\_.negex} \\
\hline

\texttt{medspacy} & Clinical NLP pipeline for medical entities, sections, and context modifiers. & \texttt{medspacy.load}, \texttt{doc.ents}, \texttt{doc.\_.sections}, \texttt{ent.\_.modifiers} \\
\hline

\texttt{vaderSentiment} & Rule-based sentiment scoring. & \texttt{SentimentIntensityAnalyzer}, \texttt{polarity\_scores} \\
\hline

\texttt{textblob} & Sentiment, subjectivity, noun phrases, POS tagging, and spelling correction. & \texttt{TextBlob.sentiment}, \texttt{noun\_phrases}, \texttt{tags}, \texttt{correct} \\
\hline

\texttt{better\_profanity} & Lightweight profanity detection, censoring, and custom bad-word loading. & \texttt{contains\_profanity}, \texttt{censor}, \texttt{load\_censor\_words} \\
\hline

\texttt{nrclex} & Emotion classification using NRC emotion lexicons. & \texttt{NRCLex}, \texttt{raw\_emotion\_scores} \\
\hline

\texttt{textstat} & Readability and text complexity metrics. & \texttt{flesch\_reading\_ease}, \texttt{flesch\_kincaid\_grade}, \texttt{smog\_index}, \texttt{sentence\_count} \\
\hline

\texttt{readability} & Additional readability metrics from \texttt{py-readability-metrics}. & \texttt{Readability}, \texttt{gunning\_fog}, \texttt{dale\_chall}, \texttt{coleman\_liau}, \texttt{ari} \\
\hline

\texttt{language\_tool\_python} & Grammar and language-error checking. & \texttt{LanguageTool}, \texttt{check} \\
\hline

\texttt{wordfreq} & Word-frequency and Zipf-scale estimates. & \texttt{zipf\_frequency} \\
\hline

\end{tabularx}
\caption{Third-party libraries and tools allowed for rubric evaluation}
\end{table*}

\end{document}